\documentclass{article}
\usepackage{ijcai25}

\usepackage{times}
\usepackage{soul}
\usepackage{url}
\usepackage[hidelinks]{hyperref}
\usepackage[utf8]{inputenc}
\usepackage[small]{caption}
\usepackage{graphicx}
\usepackage{amsmath}
\usepackage{amsthm}
\usepackage{booktabs}
\usepackage{algorithm}
\usepackage{algorithmic}
\usepackage[switch]{lineno}
\usepackage{pdfcomment}
\usepackage{amssymb}
\usepackage{multirow}
\usepackage{xcolor}

\title{TULIP: Adapting Open-Source Large Language Models for Underrepresented Languages and Specialized Financial Tasks}

\author{
İrem Demirtaş$^1$
\and
Burak Payzun$^1$
Seçil Arslan$^1$
\affiliations
$^1$Data Science \& AI, Prometeia SPA\\
\emails
\{irem.demirtas, burak.payzun, secil.arslan\}@prometeia.com
}
\begin{document}

\maketitle

\begin{abstract}
    Thanks to the growing popularity of large language models over the years, there is great potential for their applications in finance. Despite the exceptional performance of larger proprietary models, which are presented as black-box solutions through APIs, smaller models that can be hosted on-premise present opportunities for adaptability and privacy. Especially in cases where the management of sensitive information and application of domain knowledge is important, like finance, enhancing the capabilities of smaller models becomes crucial, notably for underrepresented languages.  In this work, we introduce TULIP models, which adapt Llama 3.1 8B and Qwen 2.5 7B for domain and language adaptation, focusing on financial Turkish use cases. 
    The five-stage development pipeline involves data collection, continual pre-training (CPT), benchmark design, synthetic data generation and supervised fine-tuning (SFT).  The results show that the capabilities of the models can be enhanced to effectively accomplish targeted tasks in this specific domain and language. 
\end{abstract}
\section{Introduction} 

Large language models (LLM) changed the landscape for natural language processing with their ability to handle complex text inputs while accomplishing diverse tasks and understanding multiple languages and domains  simultaneously. Although this brought great opportunities for applications in specialized areas like finance, due to the domain-specific restrictions certain limitations also emerged.  

The imperative to overcome the linguistic and contextual challenges, especially for underrepresented languages, underscores the critical role of domain-specific adaptation. Such adaptation is fundamental to achieving the high levels of confidence and accuracy required for deploying Natural Language Processing (NLP) systems in specialized, high-stakes sectors like finance. Once adapted, these models can be applied in the use cases described below with high confidence and accuracy.

In finance, many of the back-office and front-office operations heavily rely on text processing, intelligent document understanding, and conversational AI systems. Financial institutions frequently need solutions that can analyze extensive documentation and provide immediate responses to queries from relationship managers and customers. The domain-specific nature of financial tasks—such as automating credit decisions, ensuring regulatory compliance, and developing virtual agents for lending or underwriting—presents unique challenges for natural language processing systems. While instruction tuning approaches have shown promise for many applications, they are typically optimized for shorter inputs, which significantly constrains their effectiveness for specialized financial applications, particularly in underrepresented languages where domain adaptation is crucial.

By its nature, there is an abundance of document/text-based processes in banking and it is possible to automatize and digitize those cumbersome processes to improve efficiency, service levels and reduce manual workforce including both back-office and front-office.

\begin{itemize}
    \item \textbf{Back-office:} Customer orders are processed in a petition-based manner from various channel such as fax, e-mail, scanned and image-based documents. As the solution implemented by \cite{oral2020information} shows, in the customer order processing setting, first the process type is detected and the customer orders can then be extracted by deep learning models. A mid-size bank in Turkey receives nearly 100,000 such documents and an LLM-based solution can manage over 70\% of straight-through processes (STP), which comes with 20\% manual workforce saving. 

    \item \textbf{Collecting Intelligence:} Banks have dedicated teams that read sources like Trade Registry Gazette (TRG) to craft intelligence reports. The gazette, for example, includes intelligence such as the financial status of the company, address changes, changes in management, paid-in capital structure changes, which serve as critical indicators for early-warning systems and probability of default predictions. The gazette is published daily and a specialized team follows the newspaper and convert unstructured intelligence into structured data manually. With the help of LLMs, it becomes feasible to process those documents and extract required entities, relations and store in a historical intelligence database.

    \item \textbf{Chatbots and Customer Experience:} Tier-1 banks offer a highly digitalized user experience where customers can directly interact with the bank via bots on mobile apps or platforms such as WhatsApp and Facebook. A tier-1 bank's mobile app attracts \~5 million daily users, corresponding to ~20 million daily visits and customers can carry out over 700 banking transactions. These apps usually have a built-in chatbot that assists users. With a 10\% interaction rate from these visitors, bots can handle up to 2 million daily interactions. The typical RAG approach is applied to answer users' questions from documents within the bank and around 60\% of the questions can be fetched from the documents. Connecting internal procedures, guidelines or knowledge-base documents to a bot requires scale and effort which can only be covered by a bot. As bots perform better, the call center traffic is diminished and redirects itself to bots. Through the use of bots services can be offered 24/7 and customer satisfaction can be improved.

    \item \textbf{Financial Analysis:} One of the core functions of banks is credit lending for customers in various segments such as small-medium enterprises (SME), corporations and commercial. Credit lending is a document-heavy process produces a credit analysis report using inputs such as documents including financial statements of the company, intelligence sources (such as TRG), sectoral reports and previous credit analysis reports. Credit analysis report of a company can be generated using a domain-specific LLM, which can utilize nuanced understanding of financial terms required for understanding. This manual process is heavily regulated and takes around 5-7 days for a solo-company and a month for a consolidated group analysis. While the creation of the report is not a new service, automation powered by LLMs can significantly accelerate the process, enhancing efficiency and reducing manual effort.
\end{itemize}

Despite these models' high-quality generation abilities, their reliability remains a question, especially when the use of domain knowledge is a concern. Many proprietary and open-source models of various sizes are offered, yet assessing their applicability to financial problems and steering their behavior towards desirable outputs is a challenge.  

Proprietary and large open-source models are shown to offer a better performance overall, making them applicable for domains like finance, but these models come with various challenges. Due to the confidentiality and sensitivity of the data  in certain financial applications, developers may hesitate to use these models or companies may outright forbid them from transmitting any crucial information over external APIs. In other cases where an open-source model can be hosted on-prem, running and maintaining gets harder as the model size increases. These make smaller, domain-specific models that provide comparable performance to larger or proprietary models more appealing.  

Furthermore, although LLMs are pre-trained with large amounts of data fetched mostly from the internet, the limited presence of underrepresented languages like Turkish hinders their proficiency. This is more evident in smaller models, which are already limited compared to larger models due to their number of parameters. Particularly in domain adaptation in underrepresented languages, continual pre-training and post-training remain preferred solutions. Domain-specific adaptation is particularly crucial for enhancing the performance of Language Models (LLMs) in underrepresented languages, such as Turkish. The following inherent linguistic characteristics and domain-specific complexities highlight why such targeted adaptation is necessary and ultimately yields significant benefits:
\begin{itemize}
    \item \textbf{Morphology:} Morphological qualities like agglutination bring different mechanics, which may amplify performance discrepancies which also differ in terms of representation in training data. This is especially evident in smaller models.
    \item \textbf{Grammar:} The more grammatical structures are shared among highly represented European languages like English, French, Spanish, Italian, which is not the case for Turkic languages.
    \item \textbf{Language Diffusion:} Historical influence of languages from the same region result in a diverse vocabulary. This is prevalent in Turkish, which has many loanwords from languages from completely different language families (like English, French, Arabic and Persian). This can be observed even more frequently in specific domains. For example, in finance, Turkish has a catalog of terms that come from these languages and are only used in this domain. 
    \item \textbf{Domain-Specific Terminology:} Especially in highly regulated domains like finance, specific terminology comes from regulations and guidelines derived from EU/US custom terminology which creates extraordinary complexity in language understanding even for a human.
    \item \textbf{Performance Discrepancy:} An LLM with a high performance in general domains in an underrepresented language may fail to show the same performance in accuracy or understanding in complicated domains. This is underlined by the development of domain-specific models for both large and small language models alike, such as Med-Gemini \cite{medgemini} and AdaptLLM \cite{adaptllm}.
\end{itemize}


In this work, the following contributions are presented:

\begin{itemize} 
    \item Application of continual pre-training and supervised fine-tuning for domain knowledge acquisition in finance in an underrepresented language, 
    \item Task-specific data synthesis to produce supervised fine-tuning datasets, 
    \item Development of two new datasets (one to be shared with the community) in finance tasks that correspond to real-life applications sought by financial institutions,
    \item Performance evaluation on real-life finance tasks 
\end{itemize} 

\section{Related Work}

Over the years following the release of general purpose LLMs, adaptations for the domain of finance were highly popularized. Bloomberg was one of the first contenders with BloombergGPT, their 50B-parameters model trained on a mix of internal and public data, which outperforms its peers \cite{wu2023bloomberggpt}. FinGPT describes a holistic approach to developing financial LLMs from data collection to applications. The work entailed the development LoRA-finetuned models for low-cost adaptation as well \cite{fingpt}. The authors follow up this work with Instruct-FinGPT in which they fine-tune Llama 7B for sentiment analysis using open-source datasets and report better performance compared to previous financial models and Llama 7B \cite{instruct-fingpt}. Microsoft announced AdaptLLM series for adapting a 7B-parameter Llama model for domain-specific LLMs in three domains including finance and outperformed out-of-the-box models of various by training on a synthetically generated instruction dataset \cite{adaptllm}. \cite{bhatia-etal-2024-fintral} apply continual pre-training, supervised finetuning and DPO on Mistral 7B using a large-scale dataset they collected online (FinSet), existing instruction datasets and AI feedback respectively, to produce a financial LLM. FinGEITje is a financial LLM developed for the underrepresented Dutch language. The model is a finetune of Mistral 7B on translated data, which is shown to be a viable approach for customizing smaller models despite possible mistranslations \cite{dutch_financial_llm}.

For Turkish adaptation, \cite{turkish-ytu} present a comprehensive study on corpus selection for adaptation of LLMs to Turkish using both synthetically generated datasets and those translated from English and compare GPT-2 and Llama 8B variants to show that superior performance can be attained compared to other adapted Turkish LLMs. \cite{toraman-2024-adapting} also applies continual pre-training and instruction tuning on both translated and manually annotated datasets and performs an ablation study to show that the best performance is achieved when instruction turning and task-specific fine-tuning are applied collectively.

\cite{shah-etal-2022-flue} create the FLUE (Financial Language Understanding Evaluation) Dataset specifically designed for NLP problems in finance and \cite{zhu-etal-2024-benchmarking} later extend it to Chinese with CFLUE (Chinese FLUE). \cite{xie2023pixiu} come up with the PIXIU framework for finance, which spans many model development stages from data collection to model training, with an emphasis on using open-sources, covering multiple tasks and modalities to ensure diversity. Similarly, \cite{xie2024FinBen}  announce the Open FinLLM benchmark, which uses tens of datasets that cover English and Spanish languages, although there appears to be some data quality issues and the problems are not always applicable to real-life use-cases. 


\section{Data Collection} 
In domain adaptation, data collection is crucial for ensuring a model effectively processes domain-specific content. For low-resource languages like Turkish in finance, diverse and high-quality data is essential for a robust language model. This section outlines the objectives, strategies, and sources used for data collection in continual pre-training.

During the preprocessing phase, efforts were made to filter and curate high-quality, domain-specific content. This involved implementing processing pipelines for different data types, cleaning textual data by removing irrelevant sections, standardizing formats, and structuring the data to facilitate efficient training. 

For the supervised fine-tuning phase, synthetic datasets were generated using the aforementioned data and utilizing a more capable LLM. The data consisted of context-question-answer pairs designed to cover common tasks, such as multiple choice, summarization, and fill-in-the-blank tasks. These datasets were created to facilitate the adaptation of the model to specific financial tasks, ensuring that the model could respond appropriately to a variety of finance-related inquiries. 

\renewcommand{\arraystretch}{1.2}
\setlength{\tabcolsep}{5pt}

\subsection{Synthetic Data Generation} 

As continual pre-training on its own does not teach the model to follow certain formats, to make the model produce outputs in specific formats to accomplish selected tasks, supervised fine-tuning is required. Since the models were being developed with a certain domain and language in mind, initially a suitable open-source dataset was sought. However, meeting both requirements is impossible as the number of openly available financial text datasets in Turkish are limited and not suitable for real life application. To ensure paralellism with existing work in English, we looked into translating popular benchmark datasets from English, however automatic translation yielded low-quality translations due to the heavy use of specialized terms in the finance domain. The translation systems either opted for word-for-word translations of terms or US-centric/English-centric terms did not have an equivalent in Turkish which resulted in arbitrary translations. Mistranslations were also non-negligible where the translation system failed to match the correct Turkish financial terms. The high cost of human translation was also prohibitive, leading to the choice of an alternative to translating existing datasets. 


Thus, synthetic data generation was applied to build a supervised fine-tuning dataset that fits our requirements. Following a similar approach to AdaptLLM \cite{adaptllm} and Cosmopedia \cite{cosmopedia}, a supervised fine-tuning dataset was synthetically generated. As the initial text data was collected with real-life use-cases in mind, the tasks targeted by supervised fine-tuning were also selected to reflect those needs. Thus, raw text from four of the sources, Academic, Central Bank, News and Trade Registry Gazette, were used to create the dataset. In our approach, for each of these data sources, targeted tasks were selected. For each task, seed prompts were generated by incorporating randomly selected text chunks as reference with each seed prompt urging the model to accomplish the task using the reference text. A main prompt that asks the model to rephrase the seed prompt and answer it using the reference text was constructed and sent to GPT-4o. Using GPT-4o's structured output capability, the rephrased prompt and the produced answer were collected. Quality checks were implemented both for the rephrased prompt and the answer, and samples that did not meet these requirements were left out. Examples for the quality checks include, length checks, empty response checks and answer format checks. Answer format was especially crucial in cases where the answers were categorical.  

\subsection{Continual Pre-training (CPT)}
For continual pre-training, we selected Qwen2.5 and LLaMA 3.1 as our base models, as they already exhibit some degree of Turkish language proficiency despite their limitations in advanced financial tasks. Training a model that entirely lacks fundamental knowledge of Turkish grammar would have significantly expanded the project scope. Small-scale experiments conducted with the Falcon and Mistral models revealed that the performance improvements were not satisfactory.

We set the learning rate at 2e-5. The optimizer used is \texttt{paged\_adamw\_8bit}, which is optimized for memory efficiency. The training process employs mixed-precision training with \texttt{bf16} to accelerate computations while reducing memory usage. Model checkpoints were saved periodically for evaluation. A total of approximately 2.19B tokens were processed during continual pre-training, distributed across various categories: academic sources (1.1B), financial institutions (150M), textbooks and study materials (200M), market and business data (350M), legislation and regulations (50M), and other reports and documents (340M). 

We utilized QLoRA for efficient fine-tuning, applying LoRA configurations with an alpha of 128 and a rank of 64. All linear layers, as well as the head and embedding layers, were optimized during training \cite{hu2021loralowrankadaptationlarge,dettmers2023qloraefficientfinetuningquantized}. The quantization setup included 4-bit quantization and computations performed in \texttt{torch.bfloat16}. 

The models produced after these steps are referred to as TULIP-Llama-3.1 and TULIP-Qwen2.5 for the remainder of the paper.  

\subsection{Supervised Fine-Tuning (SFT)}

For supervised fine-tuning (SFT), we adopted the same training parameters as in continual pre-training, with a learning rate of 2e-6. The model was fine-tuned on approximately 23K instruction-answer pairs. The distribution of these pairs by source and task type is presented in Table~\ref{tab:sft_sources} and Table~\ref{tab:sft_tasks}, respectively.

\begin{table}[ht]
\centering
\begin{tabular}{lr}
\toprule
\textbf{Source} & \textbf{Percentage} \\ \midrule
Academic & 54\% \\
Central Bank & 5\% \\
News & 21\% \\
Trade Registry Gazette & 20\% \\ \bottomrule
\end{tabular}
\caption{Distribution of SFT Data by Source.}
\label{tab:sft_sources}
\end{table}

\begin{table}[ht]
\centering
\begin{tabular}{lr}
\toprule
\textbf{Task} & \textbf{Percentage} \\ \midrule
Fill-in-the-blank & 19\% \\
Multi-turn QA & 6\% \\
Multiple-choice QA & 16\% \\
Summarization & 16\% \\
True-False & 14\% \\
Key Information Extraction & 1\% \\
Table Generation & 3\% \\
Sentiment Analysis & 14\% \\
List Generation & 4\% \\
Named Entity Recognition & 3\% \\
Categorization & 4\% \\ \bottomrule
\end{tabular}
\caption{Distribution of SFT Data by Task.}
\label{tab:sft_tasks}
\end{table}

The synthetically generated dataset was subject to quality checks and only samples that passed all checks were included in the training set.


All experiments were carried out on two RTX A6000 GPUs, ensuring a controlled and reproducible environment.

The resulting models from this step are referred to as TULIP-Llama-3.1-IT and TULIP-Qwen2.5-IT throughout the rest of the paper.


\section{Evaluation} 

\begin{table*}[ht]
\centering
\begin{tabular}{llrrrrrrrr}
\toprule
\textbf{Model} & \textbf{Type} & \textbf{BI} & \textbf{FTIF} & \textbf{ECO} & \textbf{FI} & \textbf{PSF} & \textbf{PFT} & \textbf{AFP} & \textbf{Mean} \\ \midrule
\textbf{Llama3.1}   &
$\bigcirc$
& 0.541                                     & 0.568                                                       & 0.586                         & 0.544                                      & 0.523                                      & 0.537                                      & 0.511                                                  & 0.544                             \\
\textbf{TULIP-Llama3.1} & $\bigstar$    & \textbf{0.575}                            & \textbf{0.612}                                              & \textbf{0.626}                & \textbf{0.573}                             & \textbf{0.577}                             & \textbf{0.591}                             & \textbf{0.531}                                         & \textbf{0.583}                    \\
\textbf{TULIP-Llama3.1-IT} & $\bigstar+\blacklozenge$ & 0.574                                     & 0.596                                                       & 0.616                         & 0.567                                      & 0.552                                      & 0.581                                      & 0.525                                                  & 0.573                             \\
\textbf{Llama3.1-Instruct} & $\square$  & 0.555                                     & 0.583                                                       & 0.602                         & 0.555                                      & 0.544                                      & 0.565                                      & 0.528                                                  & 0.562                             \\ \midrule
\textbf{Qwen2.5}   & $\bigcirc$          & 0.582                                     & 0.622                                                       & 0.639                         & 0.605                                      & 0.609                                      & 0.575                                      & 0.566                                                  & 0.600                             \\
\textbf{TULIP-Qwen2.5} & $\bigstar$     & \textbf{0.671}                            & \textbf{0.680}                                              & \textbf{0.713}                & \textbf{0.667}                             & 0.652                                      & \textbf{0.666}                             & \textbf{0.626}                                         & \textbf{0.668}                    \\
\textbf{TULIP-Qwen2.5-IT} & $\bigstar+\blacklozenge$  & 0.659                                     & 0.664                                                       & 0.699                         & 0.655                                      & \textbf{0.661}                             & 0.660                                      & 0.615                                                  & 0.659                             \\
\textbf{Qwen2.5-Instruct}  & $\square$  & 0.572                                     & 0.592                                                       & 0.618                         & 0.592                                      & 0.584                                      & 0.570                                      & 0.553                                                  & 0.583                             \\ \bottomrule
\end{tabular}
\caption{Exams benchmark results in accuracy ($\bigcirc$:Base model, $\square$:Base instruct model, $\bigstar$:Continual pre-training applied on base model, $\blacklozenge$:Supervised fine-tuning)}
\label{table:exams}
\end{table*}
\subsection{Academic Exams}
Since our goal was to provide the language model with competence in both Turkish and finance, we wanted to directly measure its proficiency using a relevant benchmark. However, as there was no pre-existing benchmark designed specifically to assess financial knowledge in Turkish, we compiled our own dataset. For this purpose, we developed FINTR-EXAMS, a multiple-choice dataset compiled from undergraduate finance courses.The quality of question-answer pairs was manually checked by multiple people.

This benchmark evaluates financial literacy in Turkish across seven specialized domains: Banking and Insurance (\textbf{BI}); Foreign Trade and International Finance (\textbf{FTIF}); Economics (\textbf{ECO}); Finance and Investment (\textbf{FI}); Private Sector Finance (\textbf{PSF}); Public Finance and Tax (\textbf{PFT}); and Accounting and Financial Reporting (\textbf{AFP}).

The dataset, evaluates financial literacy in Turkish by simulating real-world assessment conditions. The questions were sourced from undergraduate-level finance exams and structured to reflect the challenges faced by finance students. 


The dataset provides a realistic assessment tool, with a passing threshold for humans set at 60 points out of 100. By benchmarking our models against FINTR-EXAMS, we assess their ability to understand and respond accurately to finance-related questions in Turkish. The model's performance is measured based on the number of correctly answered multiple-choice questions, providing insights into its strengths and weaknesses across different financial domains. 

\begin{figure}[!ht] 
\centering 
\includegraphics[width=\linewidth]{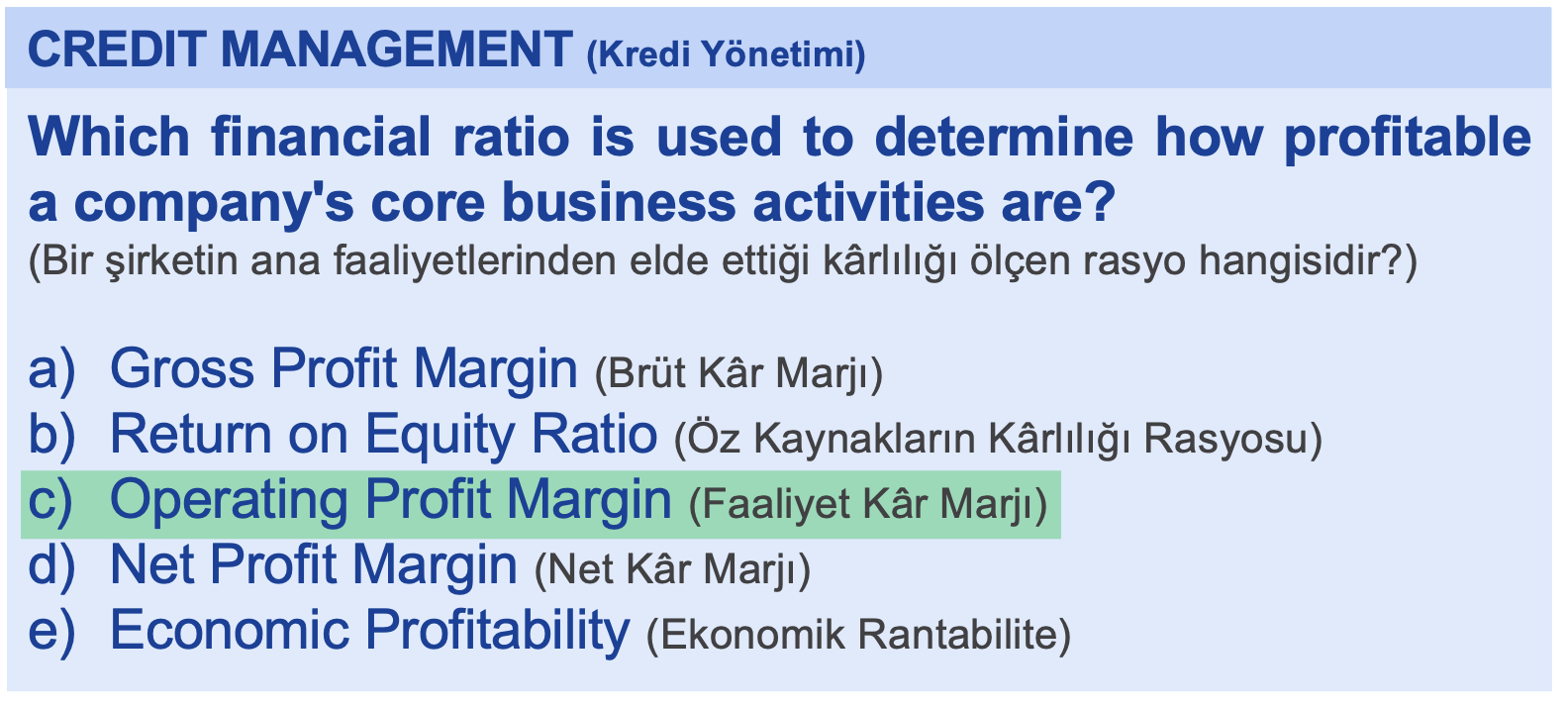} 
\caption{Question examples from the FINTR-EXAMS Benchmark.} 
\label{fig:fintr-exams-examples} 
\end{figure}

Figure \ref{fig:fintr-exams-examples} illustrates the type of questions used in the FINTR-EXAMS benchmark. \ 

Table \ref{table:exams} presents the accuracy scores of various models on the FINTR-EXAMS dataset dataset in a 5-shot setting, demonstrating their ability to comprehend and process financial knowledge in Turkish. 

The benchmark results reveal that models perform differently across financial domains due to the varying complexities of the questions. Domains like Accounting and Financial Reporting (AFP), which involve more numerical and technical questions, generally yield lower scores. This is because these questions require not only domain expertise but also a strong grasp of quantitative analysis. In contrast, domains like Banking and Insurance (BI) and Foreign Trade and International Finance (FTIF), which focus on conceptual understanding and regulatory knowledge, show better model performance.

Task-specific fine-tuning, as demonstrated by TULIP-Qwen2.5's leading performance, significantly improves model capabilities in specialized domains, contrasting with general models like Llama3.1-8B and Qwen2.5-7B-Instruct that lack such domain adaptation. In our experiments, the CPT models marginally outperformed the CPT+SFT variants on the benchmark. This suggests that CPT effectively established a strong foundational understanding of domain-specific knowledge and linguistic nuances relevant to Turkish finance, with its broader adaptation proving beneficial for benchmarks requiring generalized comprehension.

Conversely, the subsequent Supervised Fine-Tuning (SFT) process in the CPT+SFT model, using a relatively small set of examples that did not directly target the benchmark, did not yield a significant improvement and resulted in marginally lower performance. This outcome is likely attributable to the SFT phase introducing overly specific formatting constraints or adjustments that reduced the model's ability to generalize across diverse financial topics, potentially coupled with minor catastrophic forgetting or a distributional shift away from the precise competencies evaluated by the benchmark. Overall, these results indicate that while models perform well on broader conceptual topics, more specialized domains like AFP still pose a challenge, highlighting the critical need for further targeted domain adaptation.

\begin{table*}[ht]
\centering
\begin{tabular}{@{}llrrrrr @{}}
\hline
\toprule
\textbf{Model}               & \textbf{Type} & \textbf{CC} & \textbf{CM} & \textbf{CwC} & \textbf{NtC} & \textbf{Mean} \\ \midrule
\textbf{Llama3.1-Instruct}   & $\square$ & 0.881          & 0.860          & 0.846          & 0.879          & 0.866          \\
\textbf{TULIP-Llama3.1-IT}      & $\bigstar+\blacklozenge$    & \textbf{0.932} & \textbf{0.915} & 0.891          & 0.888          & 0.907          \\
\textbf{TULIP-Llama3.1-Instruct} & $\square+\blacklozenge$ & 0.881          & 0.912          & \textbf{0.926} & \textbf{0.951} & \textbf{0.918} \\ \midrule
\textbf{Qwen2.5-Instruct}    & $\square$ & 0.699          & 0.787          & \textbf{0.840} & 0.767          & 0.775          \\
\textbf{TULIP-Qwen2.5-IT}    & $\bigstar+\blacklozenge$       & \textbf{0.857} & \textbf{0.886} & \textbf{0.840} & \textbf{0.830} & \textbf{0.858} \\
\textbf{TULIP-Qwen2.5-Instruct} & $\square+\blacklozenge$ & 0.841          & 0.832          & 0.777          & 0.811          & 0.818          \\ \bottomrule
\end{tabular}
\caption{Performance comparison of models on the Turkish Trade Registry Gazette Question Answering Dataset. Accuracy measures the number of questions correctly answered, as judged by a human evaluator. ($\square$:Base instruct model, $\bigstar$:Continual pre-training applied on base model, $\blacklozenge$:Supervised fine-tuning)}
\label{table:tsg}
\end{table*}


\subsection{Turkish Trade Registry Gazette Dataset}   
Another benchmark dataset we introduce is the \textbf{Turkish Trade Registry Gazette Dataset}. The dataset contains 880 manually annotated question answer pairs that targeted at a reference text. The annotation process was carried out by 9 annotators with a Cohen's Kappa of 0.91. We share this dataset with the community through this \href{link}{link} (will be added after anonimity period).

The Turkish Trade Registry Gazette is the official source of announcements for companies in Turkey. The gazette has been publishing daily for over 70 years. The Union of Chambers and Commodity Exchanges of Turkey requires companies to announce any important trade events the a company undergoes publicly in the gazette. Events announced include foundation, change of capital, change of address, change of management, board meetings and more. The gazette follows columned newspaper layout and, without a subscription, is distributed in image PDF format. The text is extracted using Tesseract OCR and each announcement is manually labeled for one of the four event types. These events are parametrized by complex entities, relations and temporal expressions that require a layered understanding. Change of Management \textbf{(CM)} lists new assignments and removals, as well as information regarding the assignment and the persons at length. In Change of Capital \textbf{(CC)}, previous and/or final capital is announced alongside information regarding the share distibution. In Composition with Creditors \textbf{(CwC)}, the company's pre-arranged bankruptcy, defined by a term and dates, is announced. Finally, in Notice to Creditors \textbf{(NC)}, creditors of a company undergoing a merger or dissolution process are informed about how and when they can collect their debts.

The model is given an announcement that contain at least one of these types of events and is asked questions about the details. The use of language and repetitive entities of the same kind such as money amounts make these texts challenging. The model's performance is evaluated on these open-ended questions based on the number of correctly answered questions as judged by a human who marks each answer as either correct or incorrect.

The performance comparison of the models described can be found in Table \ref{table:tsg}. As depicted in the table, the additional training applied on the base models help the model outperform the instruct variants of the models. Furthermore, applying instruction tuning on the instruct variants of the models also improve the performance by around 0.05 points for both Llama 3.1 and Qwen 2.5. The benefits of continual pre-training and instruction tuning are observed especially in announcements of the type CC and CM, in which the context is longer, entities are richer and complex relations are present. For Llama, it can be seen that TULIP-Llama3.1-instruct outperforms TULIP-Llama3.1-IT in CwC and NtC. In CwC, error analysis shows that this can be attributed to the extraction of a categorical term from the text. For this term, TULIP-Llama3.1-IT and TULIP-Qwen2.5-Instruct learn to give partial answers, which invalidates their answers based on evaluation criteria. Again, in NtC, in a certain field with multiple references in the text, TULIP-Llama3.1-IT provides an explanation alongside the expected answer, which disqualifies its answer. 
An effect of training we observed in this text-generation task is that, in the cases where Qwen2.5 fails, the model usually switches languages midway while answering and outputs English or Chinese. After additional training, these behaviors are no longer observed in the outputs produced by TULIP-Qwen2.5-IT and TULIP-Qwen2.5-Instruct models.

\subsection{Efficiency of Translation}


To validate our language adaptation, we translated the FINTR-EXAMS dataset into English to assess the model's performance in English on the same problems and to explore the feasibility of a pipeline that involves translating from Turkish to English and then back into Turkish. We observed that our Turkish-adapted model not only significantly improved its performance on the Turkish set, and also demonstrated equal or better performance on the English set. This indicates that merely adding a translation layer is suboptimal, particularly when considering the potential for information loss during the back-translation of answers from English to Turkish. Detailed scores for these can be seen in Table \ref{table:translation_model_scores_formatted_detailed}.
\begin{table}[ht]
\centering
\resizebox{\columnwidth}{!}{
    \begin{tabular}{@{}lrr|r@{}}
    \toprule
    \textbf{Model}          & \textbf{Orig. (TR)} & \textbf{Self Trans. (EN)} & \textbf{GPT  Trans. (EN)} \\ \midrule
    \textbf{Llama3.1-Instruct}      & 0.550 & 0.554 & 0.633 \\
    \textbf{TULIP-Llama3.1-IT}   & \textbf{0.612} & 0.554 & 0.625 \\
    \textbf{Qwen2.5-Instruct}       & 0.604 & 0.628 & 0.711 \\
    \textbf{TULIP-Qwen2.5-IT}    & \textbf{0.693} & 0.645 & 0.712 \\ \bottomrule
    \end{tabular}%
} 
\caption{Comparison of translation model performance. The values represent accuracy scores achieved by different models. ``Self Translation (English)" means that for Qwen models, the corresponding Qwen model was used for translation from Turkish to English, and similarly for Llama models, the corresponding Llama model was used. ``GPT Translation (English)" indicates that translations were performed using GPT-4o.}
\label{table:translation_model_scores_formatted_detailed}
\end{table}
 
A core motivation for this adaptation is the practical constraint of processing sensitive data locally. Regulations in Turkey often preclude the use of cloud-based services for such data, mandating the deployment of local models. Therefore, in applicable scenarios, any translation would need to be performed by the local model itself, making its initial cross-lingual (and translation) performance limitations a critical issue to address. While an English version of the dataset translated by GPT-4o (in which both the untuned base model and our adapted model achieved a strong, comparable level of performance) served to control for translation quality, this approach is not viable for production.
 
This context underscores the necessity of our adaptation, which effectively elevates the model's Turkish performance to become comparable to its English proficiency under these operational constraints, without degrading its inherent English capabilities.

\begin{figure*}[ht!] 
\centering
\includegraphics[width=\linewidth]{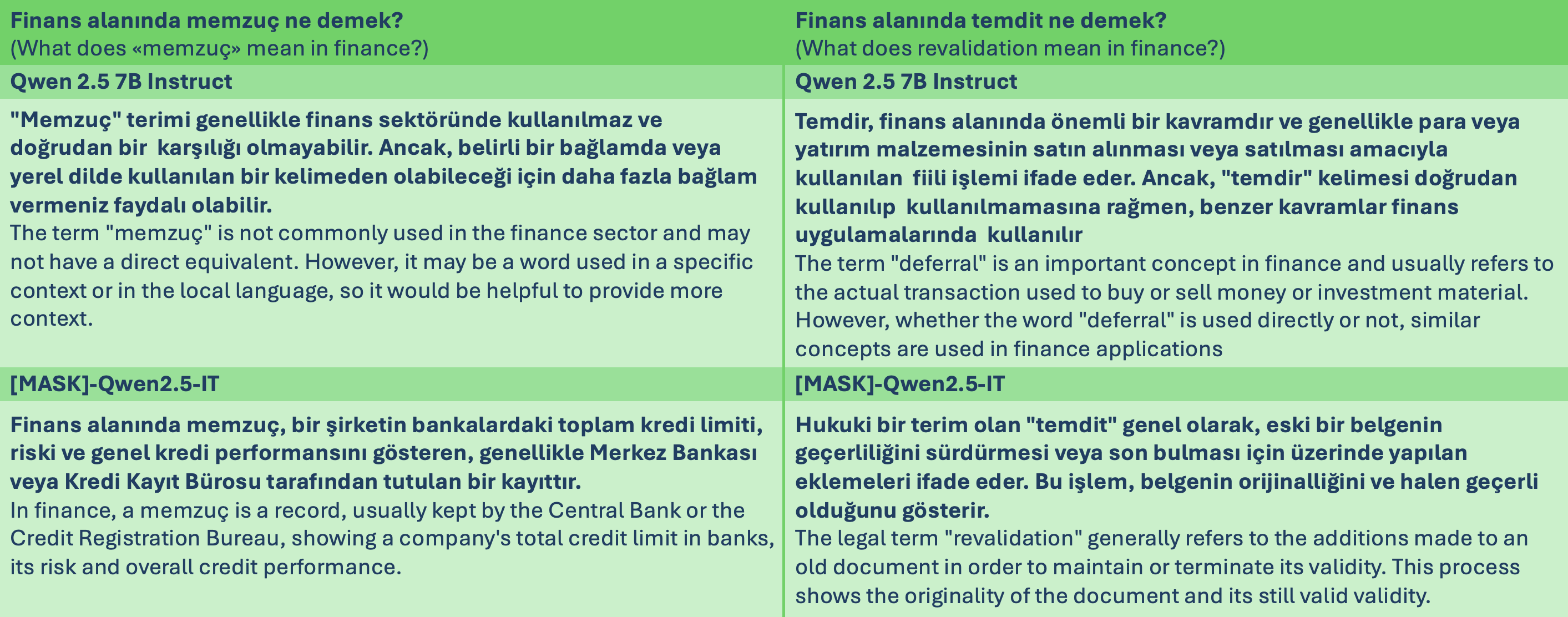} 
\caption{\textbf{Left:} The untuned model claim that ``memzuç" is not common in finance, failing to recognize its use. Our tuned model, however, accurately defines it as a consolidated record of a company's total credit limits, risk, and performance, usually kept by a central authority. \textbf{Right:} The untuned model inaccurately defines ``temdit" as ``deferral" and broadly links it to buying/selling financial assets. This misses its main meaning of ``extension" or ``rollover" of a term or contract, which our tuned model's concept of ``revalidation" (maintaining or terminating a document's validity through additions) more closely and accurately captures.} \label{fig:memzuc-example}
\end{figure*}

\subsection{Domain-Specific Financial Terminology}
Turkish financial terminology extensively incorporates loanwords from other languages, notably Persian and Arabic. Characteristically, these terms are often detached from their original, broader semantic fields in their source languages and are repurposed to denote highly specific financial applications. Furthermore, many of these terms are not commonly employed in everyday discourse.

A salient example is the Ottoman Turkish-derived term ``memzuç." While its etymological roots suggest meanings such as ``conjoined", ``mixed", ``blended", or ``unified", in financial contexts today, ``memzuç" refers to a comprehensive report detailing a firm's aggregate credit limit within the banking system. This report also includes information on existing financial issues or difficulties, the number of banking relationships, and a detailed record or statement of payments. Such information, systematically compiled and presented by the Risk Center of the Central Bank of the Republic of Turkey (CBRT), plays a critical role, particularly within credit assessment departments. Figure \ref{fig:memzuc-example} further exemplifies the specific financial usage of ``memzuç" and the misinterpretation by untuned model.

Our observations indicate that general-purpose language models, not specifically fine-tuned for the financial domain, often fail to accurately comprehend the specialized meanings of these terms. In contrast, our proposed model, which has been specifically adapted for this domain, demonstrates superior performance in the interpretation and contextual understanding of such specialized financial vocabulary.

\section{Conclusion}

This work details the successful specialization of two open-source Large Language Models (LLMs) for the finance domain in Turkish language through continual pre-training and supervised fine-tuning. Our methodology involved curating extensive financial text data, developing domain-specific benchmarks, and implementing a synthetic data generation process. The resulting models demonstrated superior performance on these benchmarks compared to their off-the-shelf counterparts and adhered well to designed task formats.

While the current task coverage offers room for expansion, this study underscores the critical role of continual pre-training in unlocking capabilities beyond what instruction tuning alone can achieve. It also validates the significant value of smaller, domain-specific models, which balance enhanced performance with practical benefits like ease of hosting, scalability, and cost-effectiveness.

These advancements are ready for impactful real-world financial applications, particularly in banking, offering data confidentiality, on-premises deployment, and manageable costs. Potential benefits include automation of back-office and front-office tasks, leading to workforce savings, increased efficiency, more straight-through processing, and improved customer satisfaction.

Future work will focus on broadening task coverage to enhance model capabilities and integrating user feedback for improved alignment with practical expectations.

\medskip
\noindent\textbf{Key Takeaways:}
\par\nobreak 
\begin{itemize}
    \item \textbf{Continual Pre-training is Crucial:} It unlocks deeper domain adaptation and new capabilities often unachievable by only fine-tuning instruct models.
    \item \textbf{Smaller, Specialized Models Offer Significant Value:} They provide a strong return on investment through domain-specific performance gains, easier deployment, and scalability, especially for focused applications.
    \item \textbf{Domain-Specific Data and Benchmarks are Essential:} Meaningful specialization and robust evaluation hinge on high-quality, relevant data and tailored benchmarks.
    \item \textbf{Bridging the Language Gap:} This customization approach can significantly improve LLM performance for underrepresented languages, such as Turkish, by instilling better term knowledge and domain-specific understanding through targeted pre-training and instruction tuning.
    \item \textbf{Practical Benefits for Industry:} Customized open-source models offer a viable path to data confidentiality, on-premises solutions, and cost-effective AI adoption in sectors like banking, automating processes and enhancing user experience.
\end{itemize}


\bibliography{anthology,custom} 

\begin{thebibliography}{}

\bibitem[\protect\citeauthoryear{Ben~Allal \bgroup \em et al.\egroup }{2024}]{cosmopedia}
Loubna Ben~Allal, Anton Lozhkov, Guilherme Penedo, Thomas Wolf, and Leandro von Werra.
\newblock Cosmopedia, 2024.

\bibitem[\protect\citeauthoryear{Bhatia \bgroup \em et al.\egroup }{2024}]{bhatia-etal-2024-fintral}
Gagan Bhatia, El~Moatez~Billah Nagoudi, Hasan Cavusoglu, and Muhammad Abdul-Mageed.
\newblock {F}in{T}ral: A family of {GPT}-4 level multimodal financial large language models.
\newblock In Lun-Wei Ku, Andre Martins, and Vivek Srikumar, editors, {\em Findings of the Association for Computational Linguistics: ACL 2024}, pages 13064--13087, Bangkok, Thailand, August 2024. Association for Computational Linguistics.

\bibitem[\protect\citeauthoryear{Cheng \bgroup \em et al.\egroup }{2023}]{adaptllm}
Daixuan Cheng, Shaohan Huang, and Furu Wei.
\newblock Adapting large language models to domains via reading comprehension.
\newblock {\em arXiv preprint arXiv:2309.09530}, 2023.

\bibitem[\protect\citeauthoryear{Dettmers \bgroup \em et al.\egroup }{2023}]{dettmers2023qloraefficientfinetuningquantized}
Tim Dettmers, Artidoro Pagnoni, Ari Holtzman, and Luke Zettlemoyer.
\newblock {QLoRA}: Efficient finetuning of quantized llms, 2023.

\bibitem[\protect\citeauthoryear{Hu \bgroup \em et al.\egroup }{2021}]{hu2021loralowrankadaptationlarge}
Edward~J. Hu, Yelong Shen, Phillip Wallis, Zeyuan Allen-Zhu, Yuanzhi Li, Shean Wang, Lu~Wang, and Weizhu Chen.
\newblock {LoRA}: Low-rank adaptation of large language models, 2021.

\bibitem[\protect\citeauthoryear{Kesgin \bgroup \em et al.\egroup }{2024}]{turkish-ytu}
H.~Toprak Kesgin, M.~Kaan Yuce, Eren Dogan, M.~Egemen Uzun, Atahan Uz, Elif İnce, Yusuf Erdem, Osama Shbib, Ahmed Zeer, and M.~Fatih Amasyali.
\newblock Optimizing large language models for turkish: New methodologies in corpus selection and training.
\newblock In {\em 2024 Innovations in Intelligent Systems and Applications Conference (ASYU)}, page 1–6. IEEE, October 2024.

\bibitem[\protect\citeauthoryear{Noels \bgroup \em et al.\egroup }{2024}]{dutch_financial_llm}
Sander Noels, Jorne De~Blaere, and Tijl De~Bie.
\newblock A dutch financial large language model.
\newblock In {\em Proceedings of the 5th ACM International Conference on AI in Finance}, ICAIF '24, page 283–291, New York, NY, USA, 2024. Association for Computing Machinery.

\bibitem[\protect\citeauthoryear{Oral \bgroup \em et al.\egroup }{2020}]{oral2020information}
Berke Oral, Erdem Emekligil, Se{\c{c}}il Arslan, and G{\"u}l{\c{s}}en Eryiǧit.
\newblock Information extraction from text intensive and visually rich banking documents.
\newblock {\em Information Processing \& Management}, 57(6):102361, 2020.

\bibitem[\protect\citeauthoryear{Saab \bgroup \em et al.\egroup }{2024}]{medgemini}
Khaled Saab, Tao Tu, Wei-Hung Weng, Ryutaro Tanno, David Stutz, Ellery Wulczyn, Fan Zhang, Tim Strother, Chunjong Park, Elahe Vedadi, et~al.
\newblock Capabilities of gemini models in medicine.
\newblock {\em arXiv preprint arXiv:2404.18416}, 2024.

\bibitem[\protect\citeauthoryear{Shah \bgroup \em et al.\egroup }{2022}]{shah-etal-2022-flue}
Raj Shah, Kunal Chawla, Dheeraj Eidnani, Agam Shah, Wendi Du, Sudheer Chava, Natraj Raman, Charese Smiley, Jiaao Chen, and Diyi Yang.
\newblock When {FLUE} meets {FLANG}: Benchmarks and large pretrained language model for financial domain.
\newblock In Yoav Goldberg, Zornitsa Kozareva, and Yue Zhang, editors, {\em Proceedings of the 2022 Conference on Empirical Methods in Natural Language Processing}, pages 2322--2335, Abu Dhabi, United Arab Emirates, December 2022. Association for Computational Linguistics.

\bibitem[\protect\citeauthoryear{Toraman}{2024}]{toraman-2024-adapting}
Cagri Toraman.
\newblock Adapting open-source generative large language models for low-resource languages: A case study for {T}urkish.
\newblock In Jonne S{\"a}lev{\"a} and Abraham Owodunni, editors, {\em Proceedings of the Fourth Workshop on Multilingual Representation Learning (MRL 2024)}, pages 30--44, Miami, Florida, USA, November 2024. Association for Computational Linguistics.

\bibitem[\protect\citeauthoryear{Wu \bgroup \em et al.\egroup }{2023}]{wu2023bloomberggpt}
Shijie Wu, Ozan Irsoy, Steven Lu, Vadim Dabravolski, Mark Dredze, Sebastian Gehrmann, Prabhanjan Kambadur, David Rosenberg, and Gideon Mann.
\newblock {BloombergGPT}: A large language model for finance.
\newblock {\em arXiv preprint arXiv:2303.17564}, 2023.

\bibitem[\protect\citeauthoryear{Xie \bgroup \em et al.\egroup }{2023}]{xie2023pixiu}
Qianqian Xie, Weiguang Han, Xiao Zhang, Yanzhao Lai, Min Peng, Alejandro Lopez-Lira, and Jimin Huang.
\newblock {PIXIU}: A large language model, instruction data and evaluation benchmark for finance, 2023.

\bibitem[\protect\citeauthoryear{Xie \bgroup \em et al.\egroup }{2024}]{xie2024FinBen}
Qianqian Xie, Weiguang Han, Zhengyu Chen, Ruoyu Xiang, Xiao Zhang, Yueru He, Mengxi Xiao, Dong Li, Yongfu Dai, Duanyu Feng, Yijing Xu, Haoqiang Kang, Ziyan Kuang, Chenhan Yuan, Kailai Yang, Zheheng Luo, Tianlin Zhang, Zhiwei Liu, Guojun Xiong, Zhiyang Deng, Yuechen Jiang, Zhiyuan Yao, Haohang Li, Yangyang Yu, Gang Hu, Jiajia Huang, Xiao-Yang Liu, Alejandro Lopez-Lira, Benyou Wang, Yanzhao Lai, Hao Wang, Min Peng, Sophia Ananiadou, and Jimin Huang.
\newblock The {FinBen}: An holistic financial benchmark for large language models, 2024.

\bibitem[\protect\citeauthoryear{Yang \bgroup \em et al.\egroup }{2023}]{fingpt}
Hongyang Yang, Xiao-Yang Liu, and Christina Dan~Wang.
\newblock Fingpt: Open-source financial large language models.
\newblock {\em FinLLM at IJCAI}, 2023.

\bibitem[\protect\citeauthoryear{Zhang \bgroup \em et al.\egroup }{2023}]{instruct-fingpt}
Boyu Zhang, Hongyang Yang, and Xiao-Yang Liu.
\newblock Instruct-fingpt: Financial sentiment analysis by instruction tuning of general-purpose large language models.
\newblock {\em arXiv preprint arXiv:2306.12659}, 2023.

\bibitem[\protect\citeauthoryear{Zhu \bgroup \em et al.\egroup }{2024}]{zhu-etal-2024-benchmarking}
Jie Zhu, Junhui Li, Yalong Wen, and Lifan Guo.
\newblock Benchmarking large language models on {CFLUE} - a {C}hinese financial language understanding evaluation dataset.
\newblock In Lun-Wei Ku, Andre Martins, and Vivek Srikumar, editors, {\em Findings of the Association for Computational Linguistics: ACL 2024}, pages 5673--5693, Bangkok, Thailand, August 2024. Association for Computational Linguistics.

\end{thebibliography}
\bibliographystyle{named}

\end{document}